\documentclass[letterpaper]{article} 
\usepackage{aaai2026}  
\usepackage{times}  
\usepackage{helvet}  
\usepackage{courier}  
\usepackage[hyphens]{url}  
\usepackage{graphicx} 
\usepackage{natbib}  
\usepackage{caption} 
\usepackage{amsmath}
\usepackage{multirow}
\usepackage{colortbl}
\usepackage{pifont}
\usepackage{xcolor}
\usepackage{booktabs}

\title{SparkUI-Parser: Enhancing GUI Perception with Robust Grounding and Parsing}
\author{
    Hongyi Jing\textsuperscript{\rm 2}\equalcontrib,
    Jiafu Chen\textsuperscript{\rm 1}\equalcontrib,
    Chen Rao\textsuperscript{\rm 1},
    Ziqiang Dang\textsuperscript{\rm 2},
    Jiajie Teng\textsuperscript{\rm 2},
    Tianyi Chu\textsuperscript{\rm 1},\\
    JunCheng Mo\textsuperscript{\rm 1},
    Shuo Fang\textsuperscript{\rm 2},
    Huaizhong Lin\textsuperscript{\rm 1},
    Rui Lv\textsuperscript{\rm 2\dag},
    Chenguang Ma\textsuperscript{\rm 2\dag},
    Lei Zhao\textsuperscript{\rm 1}\thanks{Corresponding author.}
}
\affiliations{
    \textsuperscript{\rm 1}Zhejiang University
    \textsuperscript{\rm 2}Ant Group\\
    \{jinghongyi.jhy, dijun.lr, chenguang.mcg\}@antgroup.com, \{chenjiafu, cszhl\}@zju.edu.cn

}

\usepackage{bibentry}

\begin{document}
\maketitle

\begin{abstract}
The existing Multimodal Large Language Models (MLLMs) for GUI perception have made great progress. However, the following challenges still exist in prior methods: 1) They model discrete coordinates based on text autoregressive mechanism, which results in lower grounding accuracy and slower inference speed. 2) They can only locate predefined sets of elements and are not capable of parsing the entire interface, which hampers the broad application and support for downstream tasks. To address the above issues, we propose SparkUI-Parser, a novel end-to-end framework where higher localization precision and fine-grained parsing capability of the entire interface are simultaneously achieved. Specifically, instead of using probability-based discrete modeling, we perform continuous modeling of coordinates based on a pre-trained Multimodal Large Language Model (MLLM) with an additional token router and coordinate decoder. This effectively mitigates the limitations inherent in the discrete output characteristics and the token-by-token generation process of MLLMs, consequently boosting both the accuracy and the inference speed. To further enhance robustness, a rejection mechanism based on a modified Hungarian matching algorithm is introduced, which empowers the model to identify and reject non-existent elements, thereby reducing false positives. Moreover, we present ScreenParse, a rigorously constructed benchmark to systematically assess structural perception capabilities of GUI models across diverse scenarios. Extensive experiments demonstrate that our approach consistently outperforms SOTA methods on ScreenSpot, ScreenSpot-v2, CAGUI-Grounding and ScreenParse benchmarks. The resources are available at \textcolor{blue}{\url{https://github.com/antgroup/SparkUI-Parser}}.
\end{abstract}


\begin{figure*}[t]
\centering
\includegraphics[width=\linewidth]{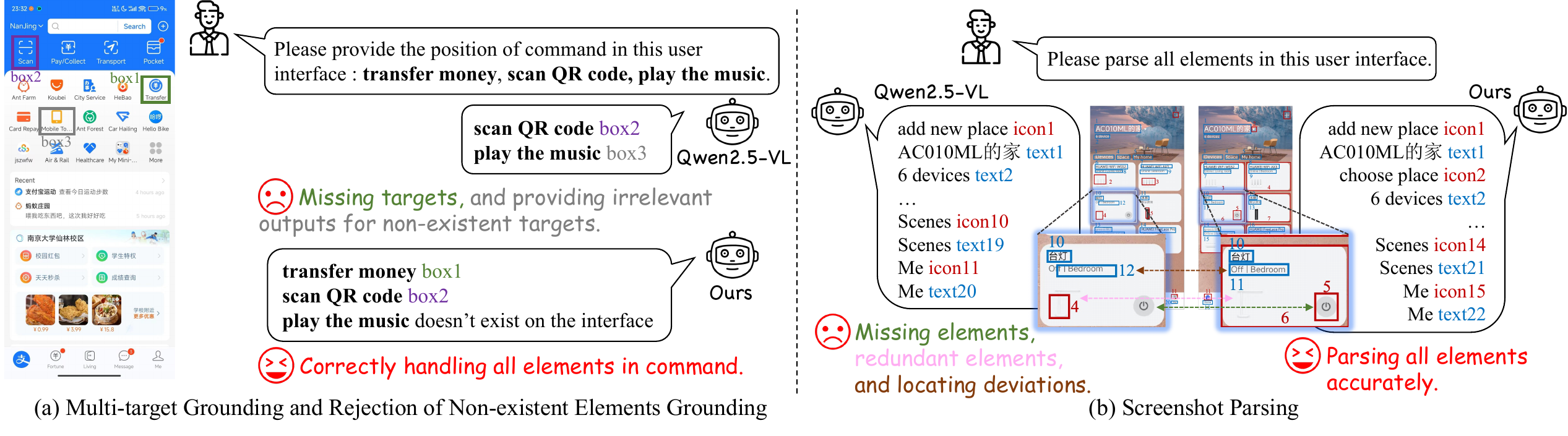}
\caption{A demonstration of multi-target grounding, rejection of non-existent elements grounding, and parsing on Qwen2.5-VL and ours.}
\label{fig:teaser}
\end{figure*}

\section{Introduction}

GUI-oriented MLLMs is capable of integrating visual and textual data to understand and interact with graphical user interfaces, providing a solid foundation for creating GUI agents. GUI agents have the potential to autonomously operate a wide range of devices, thereby transforming human-computer interaction from entirely manual process to automated and delegated workflows. Since most of the training data for general MLLMs are natural images, their perception ability on GUI images is insufficient, which limits their effectiveness in GUI-specific contexts. Natural images often contain complex scenes with diverse objects and backgrounds, whereas GUI images are more structured, featuring elements like texts, buttons, and input boxes that have specific functions and layouts.
For GUI scenario perception and advanced tasks, ideal GUI MLLMs not only need to understand the ever-changing and high information-density interfaces on various devices, but also exactly perform basic operations, such as understanding the semantics of interface elements and outputting precise coordinates.

To construct such powerful GUI MLLMs, several works \cite{cheng2024seeclick,xu2024aguvis,gou2024navigating} have already post-trained MLLMs on GUI interfaces and tasks. They focus on improving the GUI perception by locating icons or texts on GUI interfaces, which have shown promising results. However, there are still a few limitations in these previous methods in terms of perceiving GUI interfaces. 
Firstly, they are directly fine-tuned on general MLLMs, in which the text autoregressive mechanism based on the probability distribution of discrete tokens results in inaccurate results and low efficiency. 
Therefore, it may not achieve the accuracy of traditional models for grounding tasks.
Also, token-by-token prediction manner of high-precision coordinates could be a bottleneck when rapid response time is required.
Secondly, they only focus on locating and interacting with specified elements and fail to provide a detailed parsing perception of broader context and relationships between elements on the entire user interface.
Thirdly, previous models will return incorrect locations or generate irrelevant responses when a non-existent element is required to be located, leading to failures in downstream tasks, which can significantly impact user experience and system reliability.

To address the above limitations, we propose a novel end-to-end model to realize robust grounding and parsing within a model. 
Specifically, we design a route-then-predict framework to efficiently process both vision and language information, consisting of an MLLM, a token router, a vision adapter, a coordinate decoder, and an additional element matcher for training phase only. 
The output tokens from the MLLM are classified by the token router into text tokens and visual grounding tokens. Text tokens are decoded into element semantics, while visual grounding tokens, combined with visual features from the vision adapter, are processed by the coordinate decoder for localization. The lightweight coordinate decoder bridges the gap between the probability-based discrete generation of the MLLM and the continuous characteristic of image space, and provides higher-precision grounding ability and more efficient inference.
The element matcher is utilized to ensure that element semantics and coordinates are correctly matched for multi-target grounding and parsing of user interfaces.
Leveraging self-constructed parsing datasets, we train our model for parsing to extract a comprehensive representation of all elements, including their semantics and corresponding locations, empowering our model with parsing capability on the entire user interface.
Besides, we also introduce reject tokens to represent elements that do not exist in the user interface. When reject tokens appear, their coordinate decoding process is skipped straightforwardly, 
avoiding unnecessary computations and reducing the risk of generating incorrect coordinates. A demonstration of robust grounding and interface parsing is shown in Fig. \ref{fig:teaser}.

To summarize, our main contributions are listed as follow:
\begin{itemize}
    \item To the best of our knowledge, we are the first to introduce an end-to-end MLLM for GUI perception, which simultaneously achieves robust grounding and parsing on user interfaces, providing a comprehensive perception of semantics and structures.
    
    
    \item A novel route-then-predict framework by transforming discrete vocabulary coordinate modeling into continuous spatial coordinate values is proposed for better GUI grounding and parsing. By processing semantics and coordinates of the element separately, our method improves precision in grounding by around $3\%$ averagely and speeds up grounding and parsing by $5$ times and $4$ times in average.
    \item We propose a benchmark for GUI parsing, ScreenParse, which provides an evaluation for the performance of models in both locating specific elements and perceiving the overall structure of user interfaces. Moreover, metrics including element recall, element precision and semantic similarity are proposed for quantitative evaluation of parsing.
    \item The ability of multi-target grounding and rejection of non-existent elements on user interfaces are also introduced, enhancing robustness and reliability for real-world scenarios.
\end{itemize}

\begin{figure*}[t]
\centering
\includegraphics[width=0.9\linewidth]{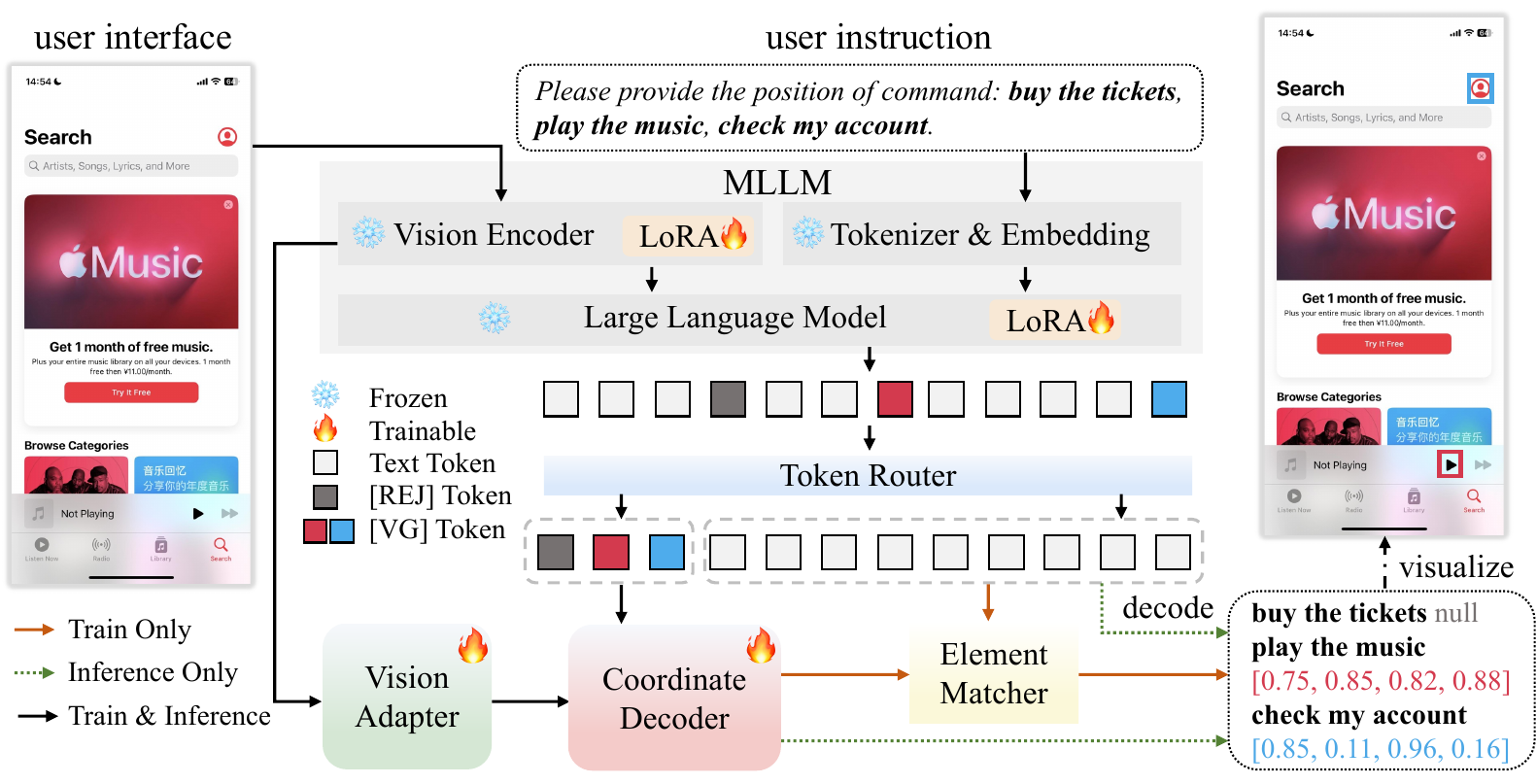}
\caption{
Overall architecture of SparkUI-Parser. A token router first classifies output tokens of the MLLM into text and special tokens ([VG] and [REJ]). Text tokens are handled by the MLLM. Features of [VG] tokens and vision-adapter features are jointly decoded for exact coordinates, while [REJ] tokens are discarded. An element matcher aligns predicted elements with ground truth during training to avoid the impact of generated elements order.
}
\label{fig:method}
\end{figure*}

\section{Related Works}
\subsection{General MLLMs}
Recent years have witnessed rapid progress in General MLLMs. GPT-4V \cite{yang2023dawn} and Gemini \cite{team2023gemini} boosted performance via massive web-scale data and reinforcement learning from human feedback, achieving strong zero-shot visual understanding and reasoning. Later, open-source models such as LLaVA \cite{liu2023llava}, Qwen-VL \cite{Qwen-VL} and InternVL \cite{chen2024internvl} leveraged instruction-tuning on curated datasets to align LLMs with visual inputs efficiently. They fused vision encoders and language decoders end-to-end, supporting diverse modalities including audio and video. Despite these advances, general MLLMs are mainly trained on natual images rather than GUI images, exhibiting limited perceptual capability in GUI-specific scenarios.

\subsection{MLLMs for GUI Perception}
To achieve GUI perception, several representative works \cite{cheng2024seeclick,gou2024navigating,wu2024atlas,yang2024aria,you2024ferret,xu2024aguvis,qin2025ui} have explored to fine-tune MLLMs with large-scale text-position pairs extracted from user interfaces. Seeclick \cite{cheng2024seeclick} achieved automatic GUI perception solely based on user interfaces for the first time, and proposed a multi-platform benchmark to evaluate GUI grounding. Ferret-UI series \cite{you2024ferret,li2024ferret} used a dynamic resolution strategy to magnify the interface details to enhance visual perception. Aguvis \cite{xu2024aguvis} and UI-TARS \cite{qin2025ui} collected a large amount of annotated data and utilized reasoning paradigms to achieve stronger GUI perception capabilities. To fully utilize the semantic and location information of elements in the user interface, OmniParser \cite{wan2024omniparser} leveraged powerful expert models to extract icons and texts, after which GPT-4V \cite{yang2023dawn} was applied to generate function of each element.  

Although they have made some progress, the following issues still exist: 1) They only achieved grounding of pre-defined elements  and could not response correctly when a non-existent element is required to be located. 2) Their way of modeling coordinates is discrete, sacrificing both grounding precision and inference speed. 3) They cannot achieve parsing on the entire interface or require the use of additional tools to complete parsing in a non-end-to-end method.
We propose an end-to-end and continuous modeling of coordinates framework to enable parsing all elements from a user interface and deal with non-existent elements correctly, providing an high-precision, efficient and effective perception of GUI interfaces.

\section{Method}

Current MLLMs-based GUI perception methods usually generate coordinates in the form of discrete text tokens. This discrete representation based on vocabulary mapping could lead to grounding inaccuracy to some extent, resulting in a performance gap compared to traditional visual grounding methods based on continuous spatial coordinate regression. To reduce the gap and improve accuracy, we design an efficient route-then-predict framework that decouples semantic understanding and coordinate optimization, while leveraging the native spatial perception capabilities of MLLMs without injecting additional perception modules. The overall architecture of SparkUI-Parser is illustrated in Fig. \ref{fig:method}. 

\subsection{Model Architecture}
The framework is composed of an MLLM, a token router, a vision adapter, a coordinate decoder, and an element matcher. Given a user instruction and a user interface, the token router classifies the output tokens of the MLLM into text tokens and special ones (including [VG] tokens and [REJ] tokens). Subsequently, multimodal features from both [VG] tokens and vision adapter are fed into the coordinate decoder for precise coordinate generation, while the [REJ] tokens are simply discarded. The text tokens are decoded by default using MLLM. During the training phase, an element matcher is employed to establish semantic and spatial correspondence between predicted and ground-truth elements, effectively mitigating the affection of output order variations. Next, we provide a detailed exposition of this process.

\subsubsection{MLLM}
Although the pre-trained MLLM have good multimodal understanding capabilities, it lacks the ability to understand and locate GUI interfaces. In order to enhance the GUI perception of MLLM, we fine-tune the vision and language part of the pre-trained MLLM using LoRA \cite{hu2022lora}.
The user instruction $T_{instruction}$ and the interface $I$ are first processed by an fine-tuned MLLM $\mathcal{F}_\mathit{MLLM}$ to obtain semantic features $f_{token}$ as
\begin{equation}
    f_{token} = \mathcal{F}_\mathit{MLLM}(I, T_{instruction}).
\end{equation}


\subsubsection{Token Router}
To model continuous coordinates based on discrete tokens, we divide the output tokens generated by MLLM into semantic tokens and location tokens through a token router. 
Based on the logits of output tokens, we identify location tokens, including [VG] tokens and [REJ] tokens. [VG] tokens refer to bounding boxes for targets, and [REJ] tokens refer to non-existent locations. The [REJ] token is later discarded, while the [VG] token requires further processing to obtain accurate coordinates for each element. The semantic tokens are decoded into natural language text via standard autoregressive generation by the MLLM.

\begin{figure}[t]
\centering
\includegraphics[width=\columnwidth]{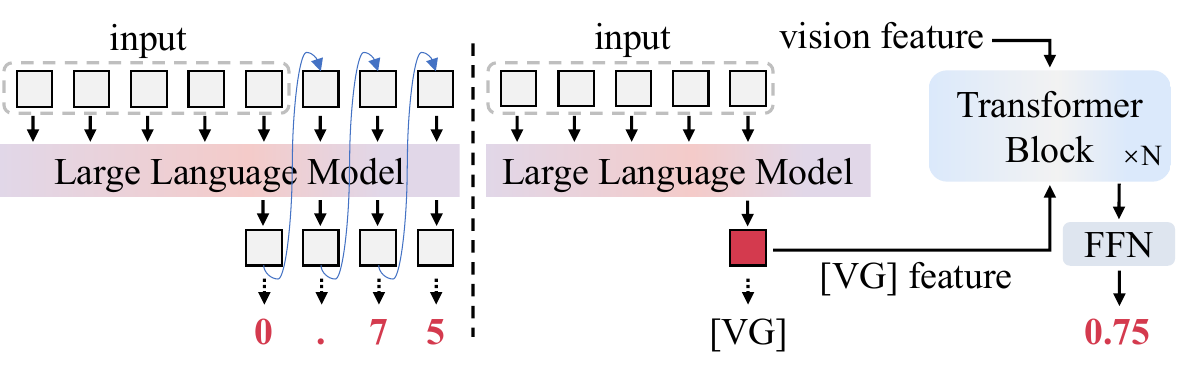}
\caption{Comparison of the coordinate generation between prior methods (left) and ours (right). 
We utilize features instead of multiple discrete tokens to obtain continuous coordinate values, thereby improving the precision of grounding and speeding up the inference. Note that only one value is displayed to simplify the visualization. In fact, a [VG] token represents coordinates of a bounding box in our method.}
\label{fig:decoder}
\end{figure}

\subsubsection{Vision Adapter}
Instead of leveraging a large-scale external vision encoder to provide visual features for accurately pinpointing the location of elements within the GUI, we utilize visual features from the vision encoder $\mathcal{F}_\mathit{ViE}$ of the MLLM and fine-tune a vision adapter $\mathcal{F}_\mathit{adapter}$ for more precise GUI-task-specific localization, which in turn outputs enhanced vision features $f_{vision}$. It can be formulated as
\begin{equation}
    f_{vision} = \mathcal{F}_\mathit{adapter}(\mathcal{F}_\mathit{ViE}(I)).
\end{equation}
This allows consistent and efficient integration of visual information with the MLLM's perception, enhancing the overall performance and adaptability of the system.

\subsubsection{Coordinate Decoder}
A lightweight coordinate decoder is introduced to enable more precise generation of continuous coordinate values. We extract the LLM's last-layer embedding corresponding to the [VG] token as text features. Along with the vision features from the vision adpater, the text features are fed into the coordinate decoder $\mathcal{F}_\mathit{deocoder}$ to obtain the precise bounding box coordinates $\mathcal{O}_\mathit{BBox}$ for each target element. The process can be formulated as
\begin{equation}
    \mathcal{O}_\mathit{BBox} = \mathcal{F}_\mathit{decoder}(f_{token_{[VG]}}, f_{vision}).
\end{equation}
The integration of visual and textual information enables the model to leverage both the semantic understanding from the text and the spatial perception from the visual features to accurately pinpoint the location of elements within the GUI. As shown in Fig. \ref{fig:decoder}, by predicting the coordinate with a transformer-based architecture, the limitation of vocabulary-based prediction of the MLLM in continuous visual grounding task is mitigated, improving grounding precision. On the other hand, only one special token is required to generate for a bounding box, compressing the tokens compared to directly output bounding box with MLLMs, thus speed up the inference speed for both grounding and parsing.

\begin{figure}[t]
\centering
\includegraphics[width=\columnwidth]{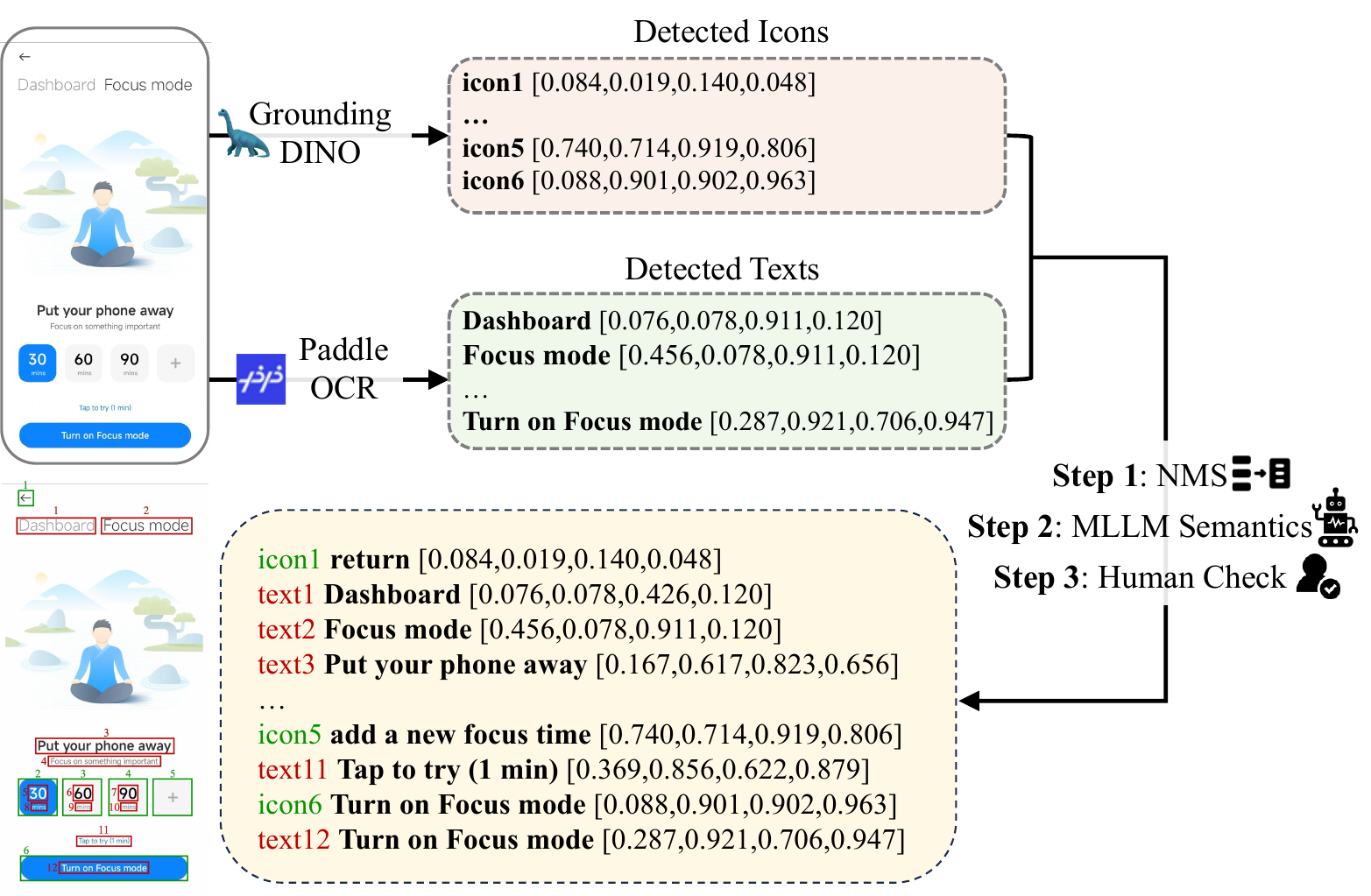}
\caption{Examples of data annotation in ScreenParse. Element types, semantics, bounding boxes for all elements are annotated.}
\label{fig:examples}
\end{figure}

\begin{table*}[t]
    \centering
    \footnotesize
    \begin{tabular}{l c c c c c c c c c}
        \toprule
        \multirow{2}{*}{\textbf{Model}} & \multirow{2}{*}{\textbf{Data}} & \multicolumn{2}{c}{\textbf{Mobile}} & \multicolumn{2}{c}{\textbf{Desktop}} & \multicolumn{2}{c}{\textbf{Web}} & \multirow{2}{*}{\textbf{Avg.}} & \multirow{2}{*}{\textbf{Time(s)}}\\
        \cmidrule(lr){3-4} \cmidrule(lr){5-6} \cmidrule(lr){7-8}
        & & \textbf{Text} & \textbf{Icon} & \textbf{Text} & \textbf{Icon} & \textbf{Text} & \textbf{Icon}\\
        \midrule
        \rowcolor{gray!20} \multicolumn{10}{l}{\textit{ScreenSpot}}\\
        GPT-4o & - & 22.6 & 24.5 & 20.2 & 11.8 & 9.2 & 8.8 & 18.3 & -\\
        Claude Computer Use & - & - & - & - & - & - & - & 83.0 & -\\
        Gemini 2.0 (Project Mariner) & - & - & - & - & - & - & - & 84.0 & -\\
        Qwen2.5-VL-7B & - & \textbf{97.1} &\underline{81.2} &86.6 &\underline{70.0} &87.4 &\underline{78.6} &\underline{84.9} & 0.763\\
        InternVL2.5-8B & - & 82.8 & 58.5 & 47.4 & 28.6 & 47.4 & 26.7 & 48.6 & 0.812 \\
        CogAgent-18B & 400K & 67.0 & 24.0 &74.2 &20.0 &70.4 &28.6 &47.4&1.112\\
        SeeClick-9.6B & 1M & 78.0 &52.0 &72.2 &30.0 &55.7 &32.5 &53.4 & 0.872\\
        OmniParser & - & 93.9 &57.0 &91.3 &63.6 &81.3 &51.0 &73.0 & -\\
        UGround-7B & 1.3M & 82.8 &60.3 &82.5 &63.6 &80.4 &70.4 &73.3& 0.821\\
        OS-Atlas-7B & 2.3M & 93.0 &72.9 &91.8 &62.9 &\textbf{90.9} &74.3 &82.5 & 0.782\\
        Aguvis-7B & 1M & \underline{95.6} &77.7 & \underline{93.8} & 67.1 & 88.3 & 75.2 & 84.4 & 0.791\\
        \textbf{SparkUI-Parser-8B} & 515K & 94.9 & \textbf{83.8} & \textbf{95.9} & \textbf{80.7} & \underline{89.6} & \textbf{82.9} & \textbf{88.0} & \textbf{0.173} \\
        \midrule
        \rowcolor{gray!20} \multicolumn{10}{l}{\textit{ScreenSpot-v2}}\\
        GPT-4o & - & 26.6 & 24.2 & 24.2 & 19.3 & 12.8 & 11.8 & 20.1 & -\\
        Qwen2.5-VL-7B & - & \textbf{99.0} & \underline{84.4} & 87.6 & 65.7 & 90.2 & \underline{79.8} & 86.5 & 0.756 \\ 
        InternVL2.5-8B & - & 84.5 & 58.3 & 48.5 & 30.0 & 43.6 & 27.1 & 48.7& 0.793\\
        CogAgent-18B & 400K & 69.3 & 27.0 & 75.8 & 20.7 & 74.4 & 31.5 & 52.8 & 1.104\\
        SeeClick-9.6B & 1M & 78.4 & 50.7 & 70.1 & 29.3 & 55.2 & 32.5 & 55.1&0.864\\
        UGround-7B & 1.3M & 84.5 & 61.6 & 85.1 & 61.4 & 84.6 & 71.9 & 76.3 & 0.805 \\
        OS-Atlas-7B & 2.3M & 95.2 & 75.8 & 90.7 & 63.6 & 90.6 & 77.3 & 84.1 & 0.774\\
        Aguvis-7B & 1M & 95.5 & 81.5 & \underline{93.3} & \textbf{77.9} & \underline{91.0} & 77.8 & \underline{87.3} & 0.779 \\
        \textbf{SparkUI-Parser-8B} & 515K & \underline{96.9} & \textbf{87.7} & \textbf{95.9} & \underline{77.1} & \textbf{93.4} & \textbf{86.2} & \textbf{89.5}& \textbf{0.168} \\
        \bottomrule
    \end{tabular}
    \caption{GUI grounding accuracy comparison of various methods on ScreenSpot and ScreenSpot-v2. In each column on different benchmarks, the \textbf{best} and \underline{second best} performance are marked, where ``-" indicates that the information cannot be obtained due to missing values or API usage.}
    \label{tab:ScreenSpot_comparison}
\end{table*}

\subsubsection{Element Matcher}
It is observed that the inherent stochasticity in textual outputs poses challenges for accurate loss computation during training due to potential element misalignment between predictions and ground truth. To mitigate this issue, an element matcher is introduced to efficiently pair each ground truth element with its closest predicted counterpart based on a predefined matching score using a modified Hungarian matching algorithm with semantics and locations, which will be stated detailedly in Sec. \ref{sec:train-obj}. This approach ensures that the model is penalized only for the differences between corresponding elements, rather than being influenced by the order in which elements are generated. This method not only simplifies the training process but also enhances the model's ability to handle diverse and complex GUI layouts, making it more adaptable to real-world scenarios.

\subsection{Training Objectives}
\label{sec:train-obj}
To train the framework on complex tasks with multiple types of instructions rather than just a simple grounding task, we elaborately design a set of objectives to ensure stable convergence during training.

The output could be multiple elements with their corresponding bounding boxes and there might exist inconsistency between the output order and ground truth order of elements. Inspired by object detection methods, DETR \cite{carion2020end}, we first compute an optimal bipartite assignment between predicted elements and ground-truth annotations using a modified Hungarian matching algorithm that jointly accounts for semantic and spatial attributes, and subsequently minimize the corresponding element-wise loss functions.

Let $y=\left\{ (t_i, b_i) \right\}$ denote the ground truth set of elements and $\hat{y}=\left\{ (\hat{t}_i, \hat{b}_i) \right\}$ the set of predictions. $t_i$ and $\hat{t}_i$ is the semantic feature of the text for the $i^{th}$ element, while $b_i$ and $\hat{b}_i$ represents its bounding box. To find a match for ground truth element $y_i$, we search the set of predictions using matching cost:
\begin{equation}
    \begin{gathered}
        \sigma = \left\{
            \begin{aligned}
                &\mathop{\arg\min}\limits_j \ \mathcal{C}_{match}(y_i, \hat{y}_j) \quad \text{if} \ \mathcal{C}_{match}(y_i, \hat{y}_j) > \mu, \\
                &\emptyset \quad \text{otherwise},
            \end{aligned}
        \right. \\
        \text{where} \ \mathcal{C}_{match} = \lambda_{IoU}\mathcal{L}_{IoU}(b_i, \hat{b}_j) + \lambda_{sem}\mathcal{L}_{sem}(t_i, \hat{t}_j).
    \end{gathered}
\label{eq:match}
\end{equation}
$\mu$ is the threshold to evaluate whether two elements match, which is experimentally set to $0.55$.

After finding a match for each ground truth element, we use cross-entropy loss for the text, $\mathcal{L}_1$ loss and IoU loss \cite{rezatofighi2019generalized} for the bounding box:
\begin{equation}
\mathcal{L}_i = \lambda_{CE}\mathcal{L}_{CE}(t_i, \hat{t}_\sigma) + \lambda_1\mathcal{L}_1(b_i, \hat{b}_\sigma) + \lambda_{IoU}\mathcal{L}_{IoU}(b_i, \hat{b}_\sigma).    
\label{eq:loss}
\end{equation}
Since each element contains semantics and coordinates optimization, for a ground truth set with $N_{gt}$ elements the total loss is defined as:
\begin{equation}
    \begin{gathered}
        \mathcal{L} = \sum_i^{N_{gt}}\mathcal{L}_i
    \end{gathered}
\end{equation}

These training objectives help with more stable learning from complex tasks with indefinite number of elements. The model can better handle the challenges posed by varying numbers of elements in different user interfaces, leading to improved performance and robustness.

\section{Benchmark}
In this section, to construct a parsing benchmark, ScreenParse, we introduce the data collection and annotation method, and provide a statistical overview of our benchmark. We also propose metrics to evaluate the performance of GUI parsing.

\subsection{Data Collection and Annotation}
We construct the parsing benchmark across different domains and applications to ensure diversity in terms of design, functionality, and complexity. 
Based on the open-source ScreenSpot benchmark \cite{cheng2024seeclick}, we re-annotated all of its interfaces. Each interface is annotated with element types, semantic labels, and bounding boxes.
To cover a broader range of user interfaces and languages, we also annotate a number of user interfaces from common Chinese applications in the same manner.

Specifically, we pre-process the user interface with expert models and manually check the quality. For all icon elements in the image, we use the open-domain detection model Grounding DINO \cite{liu2024grounding}, and we use Paddle-OCR \cite{paddleocr2020} to extract all text elements. For potentially overlapping bounding boxes, we use non-maximum suppression (NMS) for filtering. There might be some icons that still lack semantics, thus we use
a pre-trained MLLM to supplement the semantics for them. Afterwards, manually inspection will be conducted on the pre-processed data to correct any possible errors or omissions for the boxes and description information. Fig. \ref{fig:examples} illustrates the complete data-processing and annotation pipeline.

\subsection{Data Statistics}
The benchmark contains both English and Chinese user interfaces, consisting of around 800 images (400 in each language). Each interface is annotated with an average of 36 elements. A diverse set of elements is included, with common types being texts ($57.5\%$) and icons ($42.5\%$). 

\subsection{Evaluation Metrics}
Since parsing results of a user interface is evaluated for the first time, we propose to quantize the model in two aspects: bounding box and semantics performance.

\subsubsection{Bounding Box Performance}
Two metrics are introduced to evaluate the bounding box performance: \textbf{element recall} and \textbf{element precision}, which provide a comprehensive view of a model's performance in GUI parsing task. Element recall is defined as the ratio of correctly localized elements to the total number of ground truth elements, while element precision is defined as the ratio of correctly localized elements to the total number of predicted elements. 
A high element recall indicates that the model is effective at identifying the position of elements, ensuring that few actual elements are missed. A high precision indicates that most of the elements the model localizes are indeed present in the user interface, minimizing the number of incorrect grounding.

\subsubsection{Semantics Performance}
\textbf{Semantic similarity} is introduced to assess semantics performance by calculating the alignment between the model's predicted semantics and ground truth annotations for matched bounding boxes, encompassing both textual and icon description similarity. Semantic similarity evaluates the model's ability to provide semantically accurate descriptions of located elements, which is crucial for tasks demanding deep GUI understanding, such as intelligent user interaction and context-aware assistance.
\begin{table}[t]
    \centering
    \footnotesize
    \begin{tabular}{l c c c}
        \toprule
        \textbf{Model} & \textbf{Func2Box} & \textbf{Text2Box} & \textbf{Avg.}\\
        \midrule
        GPT-4o & 22.1 & 19.9 & 21.0\\
        GPT-4o with grounding & 44.3 & 44.0 & 44.2\\
        Qwen2.5-VL-7B & 59.8 & 59.3 & 59.6\\
        InternVL2.5-8B & 17.2 & 24.2 & 20.7\\
        OS-Atlas-7B & 53.6 & 60.7 & 57.2 \\
        UI-TARS-7B & 56.8 & 66.7 & 61.8 \\
        Aguvis-7B & 60.8 & \underline{76.5} & 68.7\\
        AgentCPM-GUI-8B & \underline{79.1} & \underline{76.5} & \underline{77.8}\\
        \textbf{SparkUI-Parser-8B} & \textbf{80.2} & \textbf{82.9} & \textbf{81.6} \\
        \bottomrule
    \end{tabular}
    \caption{GUI grounding accuracy comparison of various methods on CAGUI-Grounding. The \textbf{best} and \underline{second best} performance are marked.}
    \label{tab:CAGUI_comparison}
\end{table}

\begin{table*}[t]
    \centering
    \footnotesize
    \begin{tabular}{l c c c c c c c}
        \toprule
        \multirow{2}{*}{\textbf{Model}} & \multicolumn{3}{c}{\textbf{English}} & \multicolumn{3}{c}{\textbf{Chinese}} & \multirow{2}{*}{\textbf{Time(s)}} \\
        \cmidrule(lr){2-4} \cmidrule(lr){5-7}
        & \textbf{Recall} & \textbf{Precision} & \textbf{SemanticSim} & \textbf{Recall} & \textbf{Precision} & \textbf{SemanticSim} & \\
        \midrule
        GPT-4o & 5.7 & 11.7 & 0.586 & 4.9 & 7.8 & 0.406 & - \\
        Claude Computer Use & 17.1 & 35.3 & \underline{0.758} & 19.3 & 31.0 & 0.511 & - \\
        Qwen2.5-VL-7B & 18.8 & 36.6 & 0.596 & 25.0 & 52.6 & 0.854 & \underline{0.534} \\
        Qwen2.5-VL-72B & \underline{22.9} & \underline{43.7} & 0.685 & \underline{42.4} & \underline{70.5} & \underline{0.867} & 2.748 \\
        InternVL2.5-8B & 11.7 & 30.4 & 0.615 & 28.6 & 47.9 & 0.812 & 0.558 \\
        InternVL2.5-78B & 20.4 & 38.6 & 0.628 & 40.8 & 66.4 & 0.836 & 2.822 \\
        \textbf{SparkUI-Parser-8B} & \textbf{77.2} & \textbf{77.9} & \textbf{0.918} & \textbf{87.1} & \textbf{89.5} & \textbf{0.946} & \textbf{0.154} \\
        \bottomrule
    \end{tabular}
    \caption{GUI parsing evaluation of various methods on ScreenParse. The \textbf{best} and \underline{second best} performance are marked, where ``-" indicates the time consumption cannot be obtained because of API usage.}
    \label{tab:ScreenParse_comparison}
\end{table*}

\begin{table*}[t]
    \centering
    \footnotesize
    \begin{tabular}{l c c c c c}
        \toprule
        \multirow{2}{*}{\textbf{Model}} & \multicolumn{2}{c}{\textbf{ScreenSpot-v2}} & \multicolumn{3}{c}{\textbf{ScreenParse}} \\
        \cmidrule(lr){2-3} \cmidrule(lr){4-6}
        & \textbf{Text Avg.} & \textbf{Icon Avg.} & \textbf{Recall Avg.} & \textbf{Precision Avg.} & \textbf{SemanticSim Avg.} \\
        \midrule
        \textbf{full model} & \textbf{95.4} & \textbf{83.7} & \textbf{82.2} & \textbf{83.7} & \textbf{0.932}\\
        w/o coordiante decoder & 91.2 & 80.2 & 77.6 & 78.2 & 0.930\\
        w/o vision adapter & 22.1 & 17.8 & 5.4 & 4.6 & 0.915\\
        w/o element matcher & 94.7 & 83.1 & 79.5 & 80.4 & 0.814\\
        w/o parsing dataset & 95.0 & 83.2 & 20.2 & 39.2 & 0.751 \\
        \bottomrule
    \end{tabular}
    \caption{Ablation study to analyze the effectiveness of each component in our method, for conciseness and clarity of presentation, we primarily focus on results on ScreenSpot-v2 and ScreenParse.}
    \label{tab:abalation study}
\end{table*}

\section{Experiments}
\subsection{Experimental Setting}
\subsubsection{Network Architecture}
In our work, we use InternVL2.5-8B \cite{chen2024expanding} as the base MLLM $\mathcal{F}_\mathit{MLLM}$, and the vision adapter $\mathcal{F}_\mathit{adapter}$ is an MLP module with channels $[4096,2048,1024,256]$.
The coordinate decoder is a transformer-based architecture with cross-attention mechanism where the dimension of hidden states is $256$, and it finally regresses to the coordinates through a decoding head.
\subsubsection{Implementation Details}
We adopt 8 NVIDIA 80G A100 to train the whole framework, and the vision encoder is unfrozen during training. LoRA \cite{hu2022lora} is applied to fine-tune both vision and language part in MLLM, with lora-rank $r$ and lora-alpha $\alpha$ set to $16$ and $32$. We also adopt CosineLRScheduler as the learning rate scheduler, where the learning rate and warmup ratio are set to $3e^{-5}$ and $0.03$ respectively. In the training process, $\lambda_{CE}$, $\lambda_{1}$ and $\lambda_{IoU}$ in Eq. \ref{eq:loss} are set to $2.0$, $4.0$ and $1.0$ respectively, while $\lambda_{IoU}$ and $\lambda_{sem}$ in element matcher are both set to $1.0$ in Eq. \ref{eq:match}.
\subsubsection{Datasets}
Our training data is composed of open-source English datasets and self-annotated Chinese datasets, including 515k training samples in the fields of mobile, desktop and web. For English datasets, we sample 400k training samples for grounding task and additionally annotate 5k parsing samples. For Chinese datasets, we annotate 100k grounding and 10k parsing samples on our self-collected user interfaces. (See Supplementary for more datasets details.) 
Furthermore, to optimize training efficiency and enhance model robustness, we implemented a multi-target instruction paradigm, wherein each training instance randomly encompasses multiple objectives, departing from the conventional single-target approach.

\subsubsection{Baselines}
We compare our SparkUI-Parser with various baselines, including closed-source commercial models GPT-4o \cite{hurst2024gpt}, Claude Computer Use \cite{anthropic2024claude-computer-use}, Gemini 2.0 (Project Mariner) \cite{googledeepmind2024gemini2.0-project-mariner}, as well as open-source basic models Qwen2.5-VL series \cite{bai2025qwen2}, InternVL2.5 series \cite{chen2024expanding} and academic GUI models SeeClick, CogAgent \cite{hong2024cogagent}, Aguvis \cite{xu2024aguvis}, OS-Atlas \cite{wu2024atlas}, UGround \cite{gou2024navigating} and UI-TARS \cite{qin2025ui}.

\subsection{Grounding Results}
To evaluate GUI grounding capability of the models, we used the following benchmarks: 1. ScreenSpot, a GUI single-step grounding benchmark; 2. ScreenSpot-v2 \cite{wu2024atlas}, a re-annotation version based on ScreenSpot that corrects some coordinate and instruction errors; 3. CAGUI-Grounding \cite{zhang2025agentcpm}, a Chinese mobile text and function grounding evaluation benchmark. Compared to traditional models that employ text autoregression for visual localization, SparkUI-Parser utilizes a transformer-based coordinate decoder to aggregate visual and textual features and project them directly into continuous coordinate representations. As shown in Tab. \ref{tab:ScreenSpot_comparison}, our method achieves SOTA results with only a small amount of data. With the help of token router and coordinate decoder, our inference speed achieves the fastest, indicating greater potential for practical deployment across diverse application scenarios. Meanwhile, SparkUI-Parser outperforms other models on CAGUI-grouding in Tab. \ref{tab:CAGUI_comparison}. Evaluation on multiple GUI grounding benchmarks demonstrates that our method effectively supports and adapts to multilingual user interfaces and various instruction tasks.

\subsection{Parsing Results}
To evaluate the model's global perception of the semantics and coordinates of user interface elements, we use proposed benchmark ScreenParse that includes both English and Chinese user interfaces. The average time consumption of predicted elements is obtained by \textit{transformer} framework on NVIDIA 80G A100. As shown in Tab. \ref{tab:ScreenParse_comparison}, SparkUI-Parser demonstrates superior performance in parsing various user interfaces.
Commercial models as well as large-scale models can only locate a small number of elements, showing inferior recall and precision scores. Also, the semantic similarity of their located elements is low, exhibiting deviations in their semantic understanding of elements. By decoupling the semantics and coordinates of the parsing task and using an element matcher to maintain correct element optimization during training, our model achieves significant improvements on ScreenParse compared to other models, while also significantly reducing the average inference time of a single element.

\subsection{Ablation Study}
We conduct an ablation study to analyze and verify the effectiveness of each component in our method.
\subsubsection{Impact of Coordinate Decoder}
The second row of Tab. \ref{tab:abalation study} indicates that our proposed decoupled framework has different degrees of improvement compared with the autoregressive generation method of MLLM on various benchmarks. This also verifies that converting the discrete vocabulary form of coordinates into continuous spatial feature modeling promotes MLLM to better complete coordinate-related tasks.
\subsubsection{Impact of Vision Adapater}
The third row of Tab. \ref{tab:abalation study} shows that when the visual adapter is removed and only text features are used with self-attention to regress the coordinates, the accuracy greatly drops, indicating that the localization of user interface elements requires sufficient visual feature interaction.
\subsubsection{Impact of Element Matcher}
The forth row of Tab. \ref{tab:abalation study} demonstrates that removing the element matcher results in diminished parsing accuracy, primarily due to randomness of MLLM generation and target mismatches. This indicates that the element matcher plays a crucial role in aligning the model's outputs with the ground truth, reducing the focus on learning a completely consistent output order.
\subsubsection{Impact of Parsing Dataset}
The incorporation of annotated parsing dataset significantly enhances the model's capability for global structural perception. As shown in the fifth row of Tab. \ref{tab:abalation study}, the introduction of parsing dataset demonstrates substantial improvements in both structural perception and grounding precision, while slightly improves the performance of GUI grounding.

\section{Conclusion}
In this paper, we propose a route-then-predict framework that decouples the semantic and location perception of user interface elements. By leveraging a specialized coordinate decoder to map location features into a continuous coordinate space instead of generating coordinate tokens via traditional text autoregressive methods, our method achieves state-of-the-art performance on the following benchmarks while having the highest inference efficiency: ScreenSpot, ScreenSpot-v2, CAGUI-Grounding and ScreenParse. Meanwhile, a rejection mechanism is introduced in GUI scenario to avoid the model generating erroneous hallucination results and enhance the robustness and reliability. Moreover, we release a benchmark for user interface parsing to assess the model's global semantics and location perception of the user interface for furthur community research. 

\bibliography{aaai2026}

\begin{thebibliography}{26}
\providecommand{\natexlab}[1]{#1}

\bibitem[{Anthropic(2024)}]{anthropic2024claude-computer-use}
Anthropic. 2024.
\newblock Developing a computer use model.
\newblock Technical report, Anthropic.

\bibitem[{Authors(2020)}]{paddleocr2020}
Authors, P. 2020.
\newblock PaddleOCR, Awesome multilingual OCR toolkits based on PaddlePaddle.
\newblock \url{https://github.com/PaddlePaddle/PaddleOCR}.

\bibitem[{Bai et~al.(2023)Bai, Bai, Yang, Wang, Tan, Wang, Lin, Zhou, and Zhou}]{Qwen-VL}
Bai, J.; Bai, S.; Yang, S.; Wang, S.; Tan, S.; Wang, P.; Lin, J.; Zhou, C.; and Zhou, J. 2023.
\newblock Qwen-VL: A Versatile Vision-Language Model for Understanding, Localization, Text Reading, and Beyond.
\newblock \emph{arXiv preprint arXiv:2308.12966}.

\bibitem[{Bai et~al.(2025)Bai, Chen, Liu, Wang, Ge, Song, Dang, Wang, Wang, Tang et~al.}]{bai2025qwen2}
Bai, S.; Chen, K.; Liu, X.; Wang, J.; Ge, W.; Song, S.; Dang, K.; Wang, P.; Wang, S.; Tang, J.; et~al. 2025.
\newblock Qwen2. 5-vl technical report.
\newblock \emph{arXiv preprint arXiv:2502.13923}.

\bibitem[{Carion et~al.(2020)Carion, Massa, Synnaeve, Usunier, Kirillov, and Zagoruyko}]{carion2020end}
Carion, N.; Massa, F.; Synnaeve, G.; Usunier, N.; Kirillov, A.; and Zagoruyko, S. 2020.
\newblock End-to-end object detection with transformers.
\newblock In \emph{European conference on computer vision}, 213--229. Springer.

\bibitem[{Chen et~al.(2024{\natexlab{a}})Chen, Wang, Cao, Liu, Gao, Cui, Zhu, Ye, Tian, Liu et~al.}]{chen2024expanding}
Chen, Z.; Wang, W.; Cao, Y.; Liu, Y.; Gao, Z.; Cui, E.; Zhu, J.; Ye, S.; Tian, H.; Liu, Z.; et~al. 2024{\natexlab{a}}.
\newblock Expanding Performance Boundaries of Open-Source Multimodal Models with Model, Data, and Test-Time Scaling.
\newblock \emph{arXiv preprint arXiv:2412.05271}.

\bibitem[{Chen et~al.(2024{\natexlab{b}})Chen, Wu, Wang, Su, Chen, Xing, Zhong, Zhang, Zhu, Lu et~al.}]{chen2024internvl}
Chen, Z.; Wu, J.; Wang, W.; Su, W.; Chen, G.; Xing, S.; Zhong, M.; Zhang, Q.; Zhu, X.; Lu, L.; et~al. 2024{\natexlab{b}}.
\newblock Internvl: Scaling up vision foundation models and aligning for generic visual-linguistic tasks.
\newblock In \emph{Proceedings of the IEEE/CVF Conference on Computer Vision and Pattern Recognition}, 24185--24198.

\bibitem[{Cheng et~al.(2024)Cheng, Sun, Chu, Xu, Li, Zhang, and Wu}]{cheng2024seeclick}
Cheng, K.; Sun, Q.; Chu, Y.; Xu, F.; Li, Y.; Zhang, J.; and Wu, Z. 2024.
\newblock Seeclick: Harnessing gui grounding for advanced visual gui agents.
\newblock \emph{arXiv preprint arXiv:2401.10935}.

\bibitem[{GoogleDeepmind(2024)}]{googledeepmind2024gemini2.0-project-mariner}
GoogleDeepmind. 2024.
\newblock Gemini-2.0 (project mariner).
\newblock Technical report, GoogleDeepmind.

\bibitem[{Gou et~al.(2024)Gou, Wang, Zheng, Xie, Chang, Shu, Sun, and Su}]{gou2024navigating}
Gou, B.; Wang, R.; Zheng, B.; Xie, Y.; Chang, C.; Shu, Y.; Sun, H.; and Su, Y. 2024.
\newblock Navigating the digital world as humans do: Universal visual grounding for gui agents.
\newblock \emph{arXiv preprint arXiv:2410.05243}.

\bibitem[{Hong et~al.(2024)Hong, Wang, Lv, Xu, Yu, Ji, Wang, Wang, Dong, Ding et~al.}]{hong2024cogagent}
Hong, W.; Wang, W.; Lv, Q.; Xu, J.; Yu, W.; Ji, J.; Wang, Y.; Wang, Z.; Dong, Y.; Ding, M.; et~al. 2024.
\newblock Cogagent: A visual language model for gui agents.
\newblock In \emph{Proceedings of the IEEE/CVF Conference on Computer Vision and Pattern Recognition}, 14281--14290.

\bibitem[{Hu et~al.(2022)Hu, Shen, Wallis, Allen-Zhu, Li, Wang, Wang, Chen et~al.}]{hu2022lora}
Hu, E.~J.; Shen, Y.; Wallis, P.; Allen-Zhu, Z.; Li, Y.; Wang, S.; Wang, L.; Chen, W.; et~al. 2022.
\newblock Lora: Low-rank adaptation of large language models.
\newblock \emph{ICLR}, 1(2): 3.

\bibitem[{Hurst et~al.(2024)Hurst, Lerer, Goucher, Perelman, Ramesh, Clark, Ostrow, Welihinda, Hayes, Radford et~al.}]{hurst2024gpt}
Hurst, A.; Lerer, A.; Goucher, A.~P.; Perelman, A.; Ramesh, A.; Clark, A.; Ostrow, A.; Welihinda, A.; Hayes, A.; Radford, A.; et~al. 2024.
\newblock Gpt-4o system card.
\newblock \emph{arXiv preprint arXiv:2410.21276}.

\bibitem[{Li et~al.(2024)Li, You, Zhang, Feng, Agrawal, Li, Moorthy, Nichols, Yang, and Gan}]{li2024ferret}
Li, Z.; You, K.; Zhang, H.; Feng, D.; Agrawal, H.; Li, X.; Moorthy, M. P.~S.; Nichols, J.; Yang, Y.; and Gan, Z. 2024.
\newblock Ferret-ui 2: Mastering universal user interface understanding across platforms.
\newblock \emph{arXiv preprint arXiv:2410.18967}.

\bibitem[{Liu et~al.(2023)Liu, Li, Wu, and Lee}]{liu2023llava}
Liu, H.; Li, C.; Wu, Q.; and Lee, Y.~J. 2023.
\newblock Visual Instruction Tuning.

\bibitem[{Liu et~al.(2024)Liu, Zeng, Ren, Li, Zhang, Yang, Jiang, Li, Yang, Su et~al.}]{liu2024grounding}
Liu, S.; Zeng, Z.; Ren, T.; Li, F.; Zhang, H.; Yang, J.; Jiang, Q.; Li, C.; Yang, J.; Su, H.; et~al. 2024.
\newblock Grounding dino: Marrying dino with grounded pre-training for open-set object detection.
\newblock In \emph{European Conference on Computer Vision}, 38--55. Springer.

\bibitem[{Qin et~al.(2025)Qin, Ye, Fang, Wang, Liang, Tian, Zhang, Li, Li, Huang et~al.}]{qin2025ui}
Qin, Y.; Ye, Y.; Fang, J.; Wang, H.; Liang, S.; Tian, S.; Zhang, J.; Li, J.; Li, Y.; Huang, S.; et~al. 2025.
\newblock UI-TARS: Pioneering Automated GUI Interaction with Native Agents.
\newblock \emph{arXiv preprint arXiv:2501.12326}.

\bibitem[{Rezatofighi et~al.(2019)Rezatofighi, Tsoi, Gwak, Sadeghian, Reid, and Savarese}]{rezatofighi2019generalized}
Rezatofighi, H.; Tsoi, N.; Gwak, J.; Sadeghian, A.; Reid, I.; and Savarese, S. 2019.
\newblock Generalized intersection over union: A metric and a loss for bounding box regression.
\newblock In \emph{Proceedings of the IEEE/CVF conference on computer vision and pattern recognition}, 658--666.

\bibitem[{Team et~al.(2023)Team, Anil, Borgeaud, Alayrac, Yu, Soricut, Schalkwyk, Dai, Hauth, Millican et~al.}]{team2023gemini}
Team, G.; Anil, R.; Borgeaud, S.; Alayrac, J.-B.; Yu, J.; Soricut, R.; Schalkwyk, J.; Dai, A.~M.; Hauth, A.; Millican, K.; et~al. 2023.
\newblock Gemini: a family of highly capable multimodal models.
\newblock \emph{arXiv preprint arXiv:2312.11805}.

\bibitem[{Wan et~al.(2024)Wan, Song, Yu, Liu, Cheng, Huang, Bai, Yao, and Yang}]{wan2024omniparser}
Wan, J.; Song, S.; Yu, W.; Liu, Y.; Cheng, W.; Huang, F.; Bai, X.; Yao, C.; and Yang, Z. 2024.
\newblock Omniparser: A unified framework for text spotting key information extraction and table recognition.
\newblock In \emph{Proceedings of the IEEE/CVF Conference on Computer Vision and Pattern Recognition}, 15641--15653.

\bibitem[{Wu et~al.(2024)Wu, Wu, Xu, Wang, Sun, Jia, Cheng, Ding, Chen, Liang et~al.}]{wu2024atlas}
Wu, Z.; Wu, Z.; Xu, F.; Wang, Y.; Sun, Q.; Jia, C.; Cheng, K.; Ding, Z.; Chen, L.; Liang, P.~P.; et~al. 2024.
\newblock Os-atlas: A foundation action model for generalist gui agents.
\newblock \emph{arXiv preprint arXiv:2410.23218}.

\bibitem[{Xu et~al.(2024)Xu, Wang, Wang, Lu, Xie, Saha, Sahoo, Yu, and Xiong}]{xu2024aguvis}
Xu, Y.; Wang, Z.; Wang, J.; Lu, D.; Xie, T.; Saha, A.; Sahoo, D.; Yu, T.; and Xiong, C. 2024.
\newblock Aguvis: Unified pure vision agents for autonomous gui interaction.
\newblock \emph{arXiv preprint arXiv:2412.04454}.

\bibitem[{Yang et~al.(2024)Yang, Wang, Li, Luo, Chen, Huang, and Li}]{yang2024aria}
Yang, Y.; Wang, Y.; Li, D.; Luo, Z.; Chen, B.; Huang, C.; and Li, J. 2024.
\newblock Aria-UI: Visual Grounding for GUI Instructions.
\newblock \emph{arXiv preprint arXiv:2412.16256}.

\bibitem[{Yang et~al.(2023)Yang, Li, Lin, Wang, Lin, Liu, and Wang}]{yang2023dawn}
Yang, Z.; Li, L.; Lin, K.; Wang, J.; Lin, C.-C.; Liu, Z.; and Wang, L. 2023.
\newblock The dawn of lmms: Preliminary explorations with gpt-4v (ision).
\newblock \emph{arXiv preprint arXiv:2309.17421}, 9(1): 1.

\bibitem[{You et~al.(2024)You, Zhang, Schoop, Weers, Swearngin, Nichols, Yang, and Gan}]{you2024ferret}
You, K.; Zhang, H.; Schoop, E.; Weers, F.; Swearngin, A.; Nichols, J.; Yang, Y.; and Gan, Z. 2024.
\newblock Ferret-ui: Grounded mobile ui understanding with multimodal llms.
\newblock In \emph{European Conference on Computer Vision}, 240--255. Springer.

\bibitem[{Zhang et~al.(2025)Zhang, Lu, Fu, Huo, Yang, Wu, Si, Cong, Chen, Lin et~al.}]{zhang2025agentcpm}
Zhang, Z.; Lu, Y.; Fu, Y.; Huo, Y.; Yang, S.; Wu, Y.; Si, H.; Cong, X.; Chen, H.; Lin, Y.; et~al. 2025.
\newblock AgentCPM-GUI: Building Mobile-Use Agents with Reinforcement Fine-Tuning.
\newblock \emph{arXiv preprint arXiv:2506.01391}.

\end{thebibliography}

\end{document}